\documentclass{article}

\usepackage{PRIMEarxiv}

\usepackage[utf8]{inputenc} 
\usepackage[T1]{fontenc}    
\usepackage{hyperref}       
\usepackage{url}            
\usepackage{booktabs}       
\usepackage{amsfonts}       
\usepackage{nicefrac}       
\usepackage{kotex}       
\usepackage{microtype}      
\usepackage{lipsum}
\usepackage{fancyhdr}       
\usepackage{graphicx}       
\graphicspath{{media/}}     

\usepackage{subcaption}

\title{Imagine Networks}

\author{
  *Seokjun Kim$^{1}$, Jaeeun Jang$^{2}$, *Hyeoncheol Kim \\
  Department of Computer Science and Engineering \\
  Korea University \\
  \texttt{\{*melon7607, wkdwodms0779\}@naver.com *harrykim@korea.ac.kr} \\
}

\begin{document}
\maketitle

\begin{abstract}
In this paper, we introduce an imagine network that can simulate itself through artificial association networks.
Association, deduction, and memory networks are learned, and a network is created by combining the discriminator and reinforcement learning models. This model can learn various datasets or data samples generated in environments and generate new data samples.
\end{abstract}

\keywords{Artificial Association Networks \and The fourth paper \and (More experiments are in progress)}

\section{Introduction}
To date, various neural networks have appeared, and these networks are being used in various fields (classification, generation, detection, etc..).

However, there are tasks that humans can do, but neural network models can't.
Human's representative ability is imagination and creating something new.
In this study, we designed an imagine networks model based on artificial association networks.
It was designed by sequencing the structures imagined by humans. 

several abilities are needed to imagine.

(1) The first is recognition. If the information is not recognized, it can't be utilized. We need to recognize which object is what. We learn this through classification or clustering.

(2) The second is deduction.
We create a proposition and combine the propositions to create a compound proposition and solve our problems through the relationship between the main object and the other objects. 
And we need to learn what the results will be if we combine the objects in principle.

(3) The third is memory.
Humans have a memory input device called the hippocampus and it stores information in the brain.
And the memory device can take out the desired information and use it in the deduction process. 
It is possible to recall information based on experience without input, similar to when we close our eyes and cover ears.

(4) The fourth is that the choosing process which is to get the optimal reward.
This selector consists of a reinforcement learning structure such as Q-learning\cite{watkins1992q}, and it is known that reinforcement learning is a structure similar to the human brain. And if we choose another object by reinforcement models, it should be something better.

(5) the last part is the discriminator.
This discriminator is a previously learned recognition model(1) or a binary classification model. This process determines whether the sample is the desired one or not.

We will understand the principle relationship between the input object and the target object.

And this network is characterized by being a data-driven network that uses a neuro tree data structure.
Therefore, the generated data sample may be input to the association model again and used as a model for generating hierarchical and relational information.

We experimented in a reinforcement learning environment to learn this. 
And similar to simulating in our heads, we generated a sample that performs simulation in a network.

\section{Related works}
\label{sec:related_works}
\paragraph{Artificial Association Networks \cite{kim2021association}}
This study is the first study of artificial association networks(AANs).
In this study, it is possible to learn various datasets simultaneously, and this paper introduces various sensory organs and the association area where information is integrated.
instead of using a fixed sequence layer, the network learns according to the tree structure using a data structure called an neuro tree.
The data structure defined $\mathbf{AN} = \{x, t, \mathbf{A}_c, \mathbf{C}\}$, And propagation method is conducted using recursive convolution called DFC and DFD.

\paragraph{Deductive Association Networks \cite{kim2021deductive}}
This study is a model for the role of the frontal lobe and is a study to utilize information generally transmitted from the association area. Representatively, it is designed to be responsible for the ability to deduce and think.
This model uses the result of the previous proposition as input to the next proposition to combine various principles.

\paragraph{Memory Association Networks \cite{kim2021memory}}
This model is designed to store root vectors.
In this study, short-term memory is used to solve the class-imbalance problem, and long-term uses it to create distributions of objects.

\paragraph{Q-learning \cite{watkins1992q}}
This agent learns what information to take out of memory or what action to perform for the current state to get the optimal reward.
If the space is continuous, models such as DQN\cite{mnih2013playing}, SAC\cite{haarnoja2018soft} and DDPG\cite{lillicrap2015continuous} could be used.

\paragraph{GAN \cite{goodfellow2020generative}}
The main concept is that the generator generates a sample, and the discriminator determines whether the sample is generated by the generator or a training sample. 
There are some difference from the discriminator in this model, It is similar to this model that \textbf{a meta-observer exists} and judges. As a result, we get the sample we want.

\section{Imagine Networks}
\label{sec:imagine}
\begin{figure}[h]
\centering
{\includegraphics[width=0.80\textwidth]{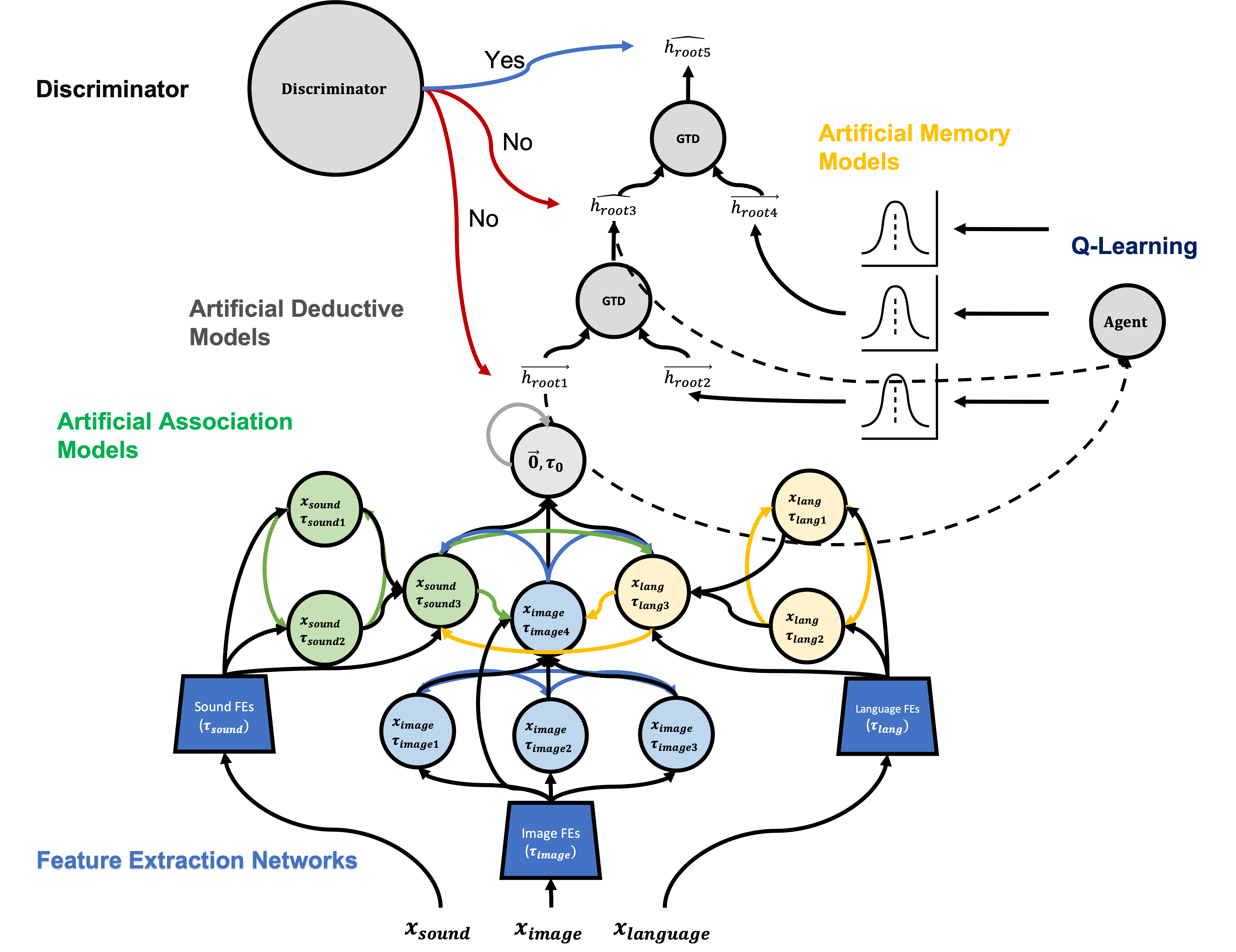}}
\caption{ Ideal Imagine Modeling }
\label{fig:long-term}
\vspace{-10px}
\end{figure}
Imagine Networks is created by combining various neural network models. This structure can be seen as the ability to perform in the frontal lobe. In addition, the neuro tree learns relational information and hierarchical information, and The generated neuro tree uses it again for an association model.
Therefore, I introduce Imagine networks, a thinking machine.

\subsection{Model 1 : Extracting the relationship between input \& target}

\begin{equation}
\label{eq:gtnn}
\overrightarrow{h}_{root1} = \mathbf{AAN}(NT)
\end{equation}
\begin{equation}
\label{eq:gtd_out}
action_{mem}, action_{op} = Agent(\overrightarrow{h}_{root1})
\end{equation}
\begin{equation}
\label{eq:gtd_out}
\overrightarrow{h}_{root2} = \mathbf{MAN}(action_{mem})
\end{equation}
\begin{equation}
\label{eqn:gtd_elements}
\mathbf{E}_{1} = (\overrightarrow{h}_{root1}, \overrightarrow{h}_{root2} ..)
\end{equation}
\begin{equation}
\label{eqn:gtd_elements}
\hat{h}_{root3} = \mathbf{DAN}(\mathbf{E}_{1}, action_{op})\label{eqn:gtdgru}
\end{equation}

\begin{equation}
\label{eqn:gtd_elements}
True/False = isPastState(\hat{h}_{root3})\label{eqn:gtdgru}
\end{equation}
\begin{equation}
\label{eqn:gtd_elements}
True/False = Discriminator(\hat{h}_{root3})\label{eqn:gtdgru}
\end{equation}
If it's not past state and not target information...

\begin{equation}
\label{eq:gtd_out}
action_{mem}, action_{op} = Agent(\hat{h}_{root3})
\end{equation}
\begin{equation}
\label{eq:gtd_out}
\overrightarrow{h}_{root4} = \mathbf{MAN}(action_{mem})
\end{equation}
\begin{equation}
\label{eqn:gtd_elements}
\mathbf{E}_{2} = (\hat{h}_{root3}, \overrightarrow{h}_{root4} ..)
\end{equation}
\begin{equation}
\label{eqn:gtd_elements}
\hat{h}_{root5} = \mathbf{DAN}(\mathbf{E}_{2}, action_{op})\label{eqn:gtdgru}
\end{equation}

\begin{equation}
\label{eqn:gtd_elements}
True/False = isPastState(\hat{h}_{root5})\label{eqn:gtdgru}
\end{equation}
\begin{equation}
\label{eqn:gtd_elements}
True/False = Discriminator(\hat{h}_{root5})\label{eqn:gtdgru}
\end{equation}
.. Repeat!

\begin{figure}[h!]
\centering
\begin{subfigure}{0.45\textwidth}
    \centering
    \includegraphics[height=3.0cm]{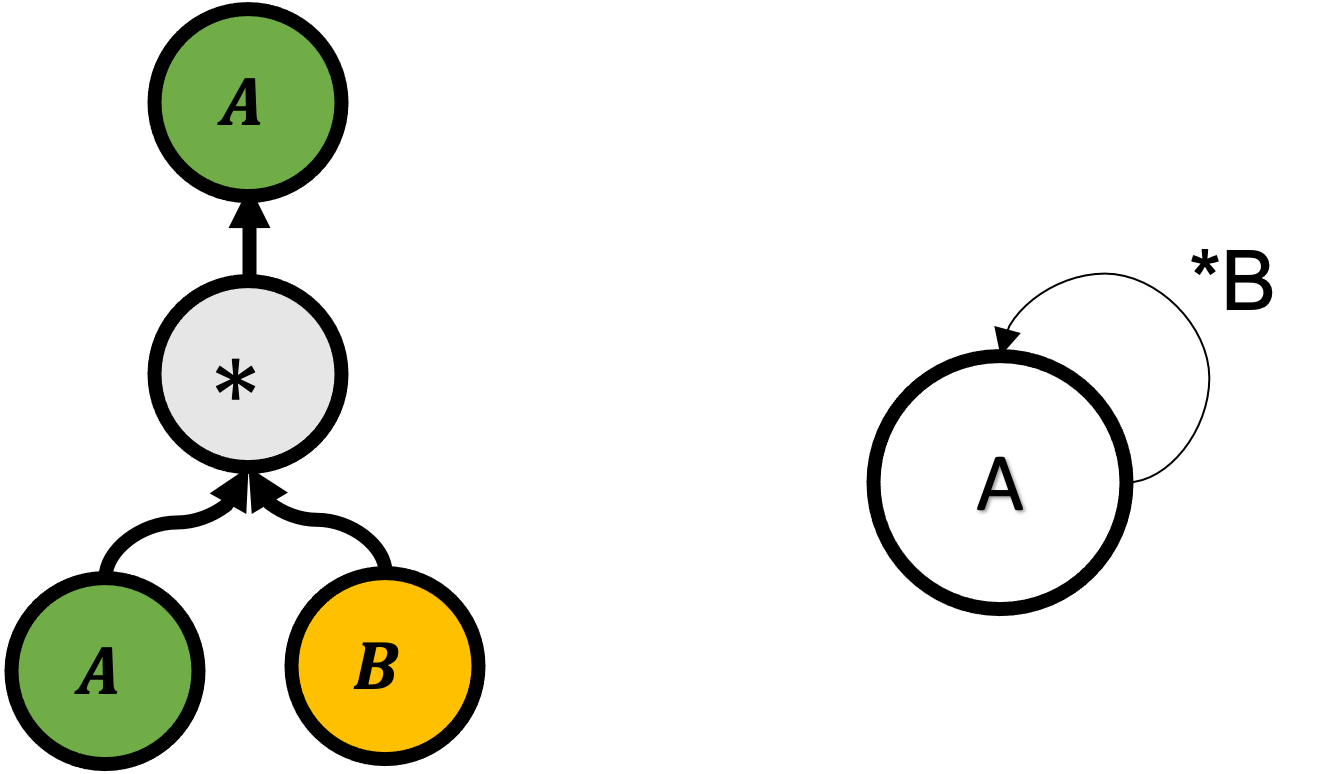}
    \caption{This case is returning to the past state}
    \label{fig:materials_first}
\end{subfigure}
\hfill
\begin{subfigure}{0.45\textwidth}
    \centering
    \includegraphics[height=3.0cm]{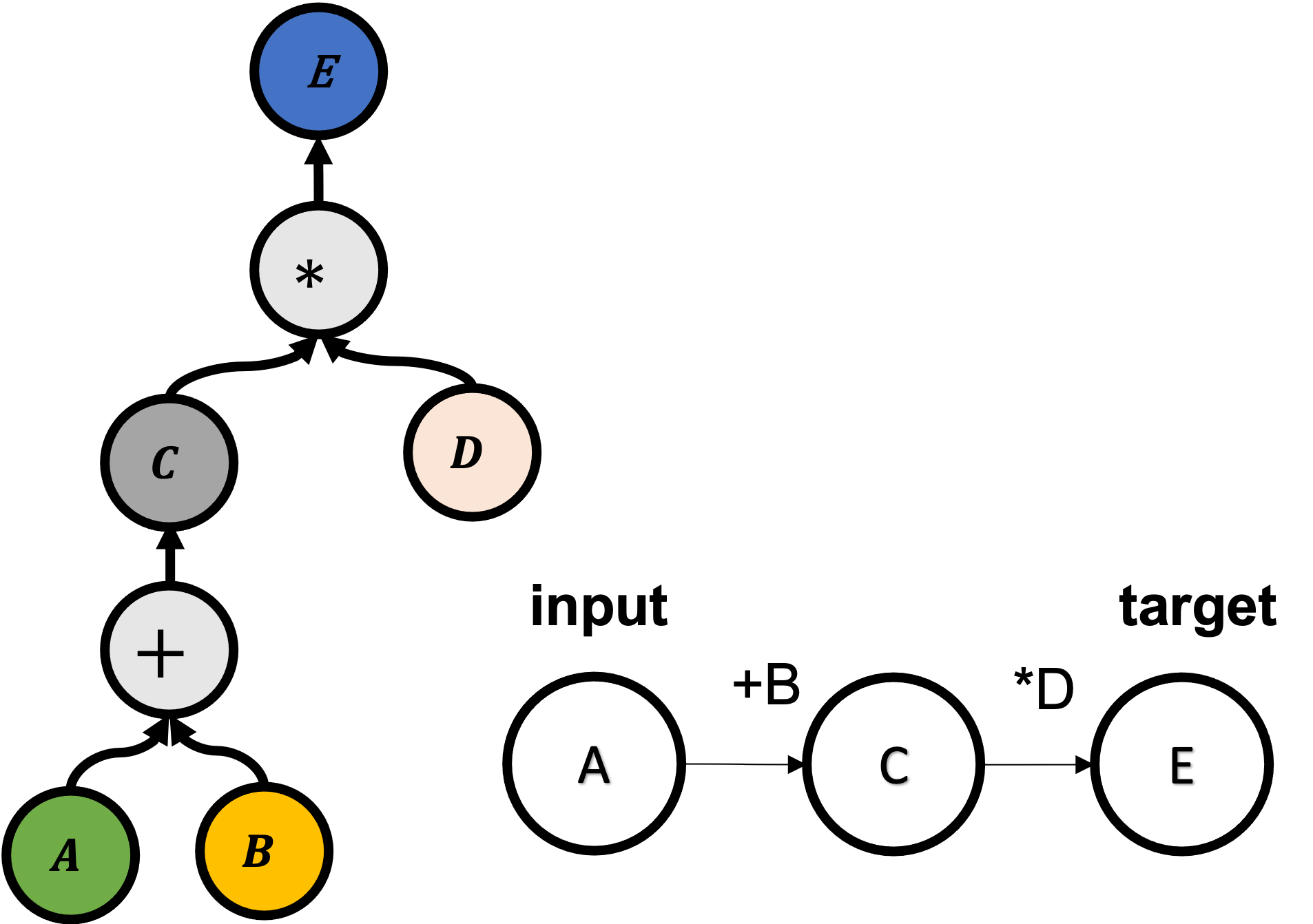}
    \caption{The relationship between input and target(goal) extracted by image networks}
    \label{fig:materials_second}
\end{subfigure}
\hfill
\caption{Extracting the relationship between input \& target}
\label{fig:materials}
\end{figure}

Fig.\ref{fig:materials_first} is the case of returning to the past state, so the value function is updated and then restarted the simulation.

Fig.\ref{fig:materials_second} represents the relationship between input and target without a recursive path.

In this way, we can use memory and computing devices to interpretively extract the relationship between input and target through reinforcement learning methods. If we store this value function, we can reuse it as an operator.

\subsection{Model 2 : Expansion of information (experiments are in progress)}
\begin{figure}[h]
\centering
{\includegraphics[width=0.70\textwidth]{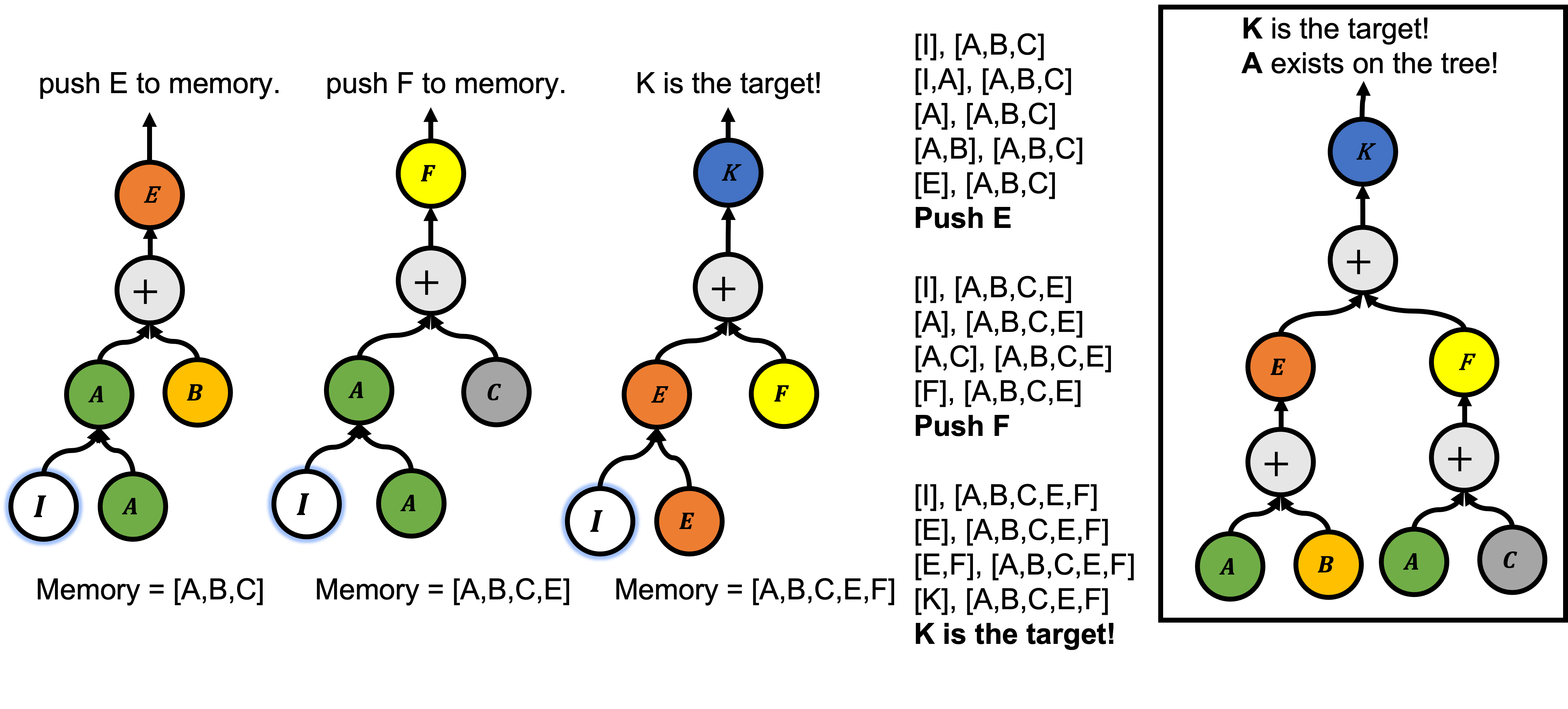}}
\caption{ how to build the imagine tree }
\label{fig:howtobuildimaginetree}
\vspace{-10px}
\end{figure}

In Model 1, there are limitations. For example, some information does not exist in memory, but information that can be generated by deductive networks cannot be utilized.
Therefore, we propose a model that can utilize the output information that starts with information from other starting point.
an identity element, which is in its initial state, is added and information can be stored in memory (Fig.\ref{fig:howtobuildimaginetree}).
\begin{figure}[h]
\centering
{\includegraphics[width=0.90\textwidth]{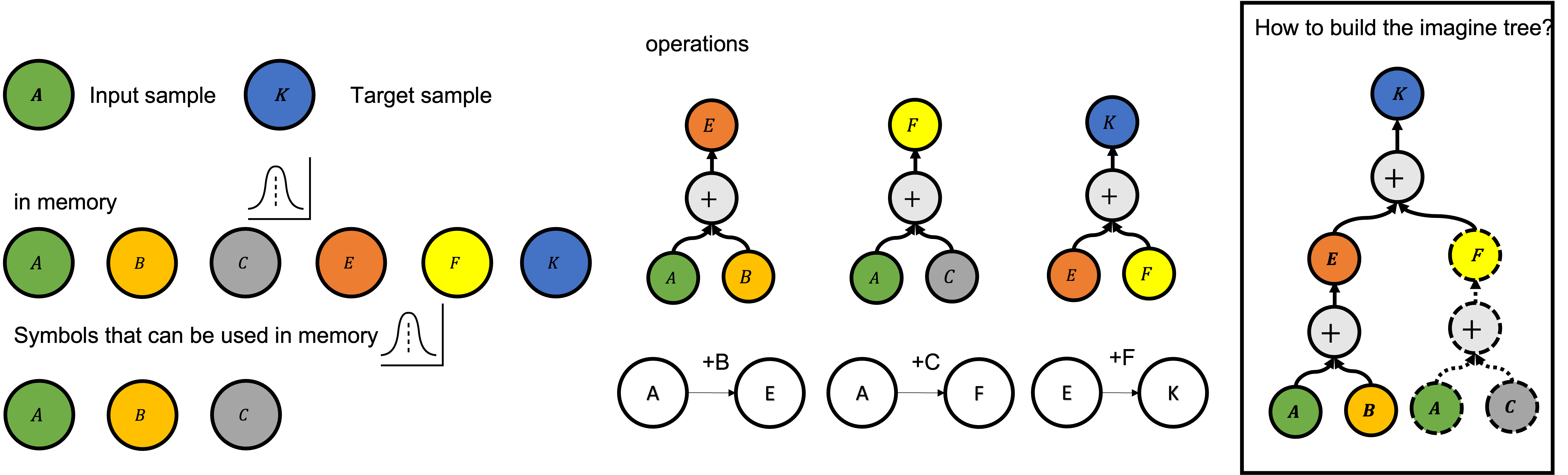}}
\caption{ the limitation of model 1 }
\label{fig:long-term}
\vspace{-10px}
\end{figure}

\begin{equation}
\label{eq:gtnn}
\overrightarrow{state}_{1} = GRU( init\_identity, \overrightarrow{state}_{0})
\end{equation}
\begin{equation}
\label{eq:gtd_out}
action_{mem} = Agent(\overrightarrow{state}_{1})
\end{equation}
\begin{equation}
\label{eq:gtd_out}
\overrightarrow{h}_{root1} = \mathbf{MAN}(action_{mem})
\end{equation}
\begin{equation}
\label{eq:gtnn}
\overrightarrow{state}_{2} = GRU( \hat{h}_{root1}, \overrightarrow{state}_{1})
\end{equation}
\begin{equation}
\label{eq:gtd_out}
action_{mem}, action_{op} = Agent(\overrightarrow{state}_{2})
\end{equation}
\begin{equation}
\label{eq:gtd_out}
\overrightarrow{h}_{root2} = \mathbf{MAN}(action_{mem})
\end{equation}
\begin{equation}
\label{eqn:gtd_elements}
\mathbf{E}_{1} = (\overrightarrow{h}_{root1}, \overrightarrow{h}_{root2} ..)
\end{equation}
\begin{equation}
\label{eqn:gtd_elements}
\hat{h}_{root3} = \mathbf{DAN}(\mathbf{E}_{1}, action_{op})\label{eqn:gtdgru}
\end{equation}

\begin{equation}
\label{eqn:gtd_elements}
True/False = isPastState(\hat{h}_{root3})\label{eqn:gtdgru}
\end{equation}
\begin{equation}
\label{eqn:gtd_elements}
True/False = isInMemory(\hat{h}_{root3})\label{eqn:gtdgru}
\end{equation}
if false .. push the output to the memory
\begin{equation}
\label{eqn:gtd_elements}
True/False = Discriminator(\hat{h}_{root3})\label{eqn:gtdgru}
\end{equation}
if not true ..

\begin{equation}
\label{eq:gtnn}
\overrightarrow{state}_{3} = GRU( init\_identity, \overrightarrow{state}_{2})
\end{equation}
\begin{equation}
\label{eq:gtd_out}
action_{mem} = Agent(\overrightarrow{state}_{3})
\end{equation}
\begin{equation}
\label{eq:gtd_out}
\overrightarrow{h}_{root4} = \mathbf{MAN}(action_{mem})
\end{equation}
\begin{equation}
\label{eq:gtnn}
\overrightarrow{state}_{4} = GRU( \overrightarrow{h}_{root4}, \overrightarrow{state}_{3})
\end{equation}
\begin{equation}
\label{eq:gtd_out}
action_{mem}, action_{op} = Agent(\overrightarrow{state}_{4})
\end{equation}
\begin{equation}
\label{eq:gtd_out}
\overrightarrow{h}_{root5} = \mathbf{MAN}(action_{mem})
\end{equation}
\begin{equation}
\label{eqn:gtd_elements}
\mathbf{E}_{2} = (\overrightarrow{h}_{root4}, \overrightarrow{h}_{root5} ..)
\end{equation}
\begin{equation}
\label{eqn:gtd_elements}
\hat{h}_{root6} = \mathbf{DAN}(\mathbf{E}_{2}, action_{op})\label{eqn:gtdgru}
\end{equation}

\begin{equation}
\label{eqn:gtd_elements}
True/False = isPastState(\hat{h}_{root6})\label{eqn:gtdgru}
\end{equation}
\begin{equation}
\label{eqn:gtd_elements}
True/False = isInMemory(\hat{h}_{root6})\label{eqn:gtdgru}
\end{equation}
if false .. push the output to the memory
\begin{equation}
\label{eqn:gtd_elements}
True/False = Discriminator(\hat{h}_{root6})\label{eqn:gtdgru}
\end{equation}
.. Repeat!

\subsection{Eureka! Learning}
\label{sec:imagine}

\paragraph{Recognition}
This process is for extracting the feature vector of the object. Classification or Clustering tasks are representative recognition learning processes in which each object can produce different feature vectors.

\paragraph{Memorization}
This process is to use the information learned in the past by recalling memories without any input. Short-term memory is stored for each class of objects, and each class has a distribution for a generation.

\paragraph{Deduction}
Deduction performs a prediction task on the proposition result that appears when the objects are combined. This is like learning a simple proposition. This step can be replaced by a style transfer model etc.

\paragraph{Reinforcement learning}
Q-learning plays the role of using currently recognized information as state and selecting it to perform deduction with other information.
And through deduction, information is combined, and optimal compensation is obtained.

\paragraph{Discriminator}
The discriminator plays a role in determining whether the currently recognized information is what I want and instructing the q-learning model to stop when the desired information comes out.

\section{Experimental results}
\label{sec:result}

\subsection{Imagination in A Reinforcement Learning Environment}
Let's apply the above description to a reinforcement learning environment.

First, We created a queue for each state in short-term memory, and samples from the environment are stored in short-term memory(state number(label), screen, action, reward, next state number(label), next screen, next action, done).

In addition, sampling was performed as much as batch-size in short-term memory, and the samples were trained to recognition, memory, deduction, discriminator, and agent model.
\subsubsection{How to train Imagine Networks}
\paragraph{Recognition} 
In the recognition process, the current screen is input and the number of the current state is recognized.
Recognition in this experiment means being aware of the current state. Since the state in this experiment is a discrete space, we numbered this state and labeled it by state, and supervised learning was performed to preset the state number by inputting the screen image of the state.
\paragraph{Memorization}
In the memorization process, the root vector shown is stored in memory networks, and the distribution is learned together with the decoder.
The memory network learns that generates a screen of each state.
Therefore, we stored image information generated from environments in short-term memory and learned the images in long-term memory. Each state's information is stored as a distribution. We can generate screen information by entering the number of the desired state.

\paragraph{Reinforcement learning}
In the agent process, the root vector means current state information. And state information is discrete space.
Therefore, we can generate q-table for each state. And encode the optimal action to be performed in the current state and use it as an input to the deductive model.

In the future, we can change this process to perform an action on "which sample should be taken out of memory".

Reinforcement learning in this experiment has randomness and serves as a role in generating data in the environment. 
A selector selects an action to move by finding the shortest path.

\paragraph{Deduction}
In the deduction process, Now, there is a root vector of the current state and action information to be performed, and there is the next screen information that appears when the action is performed. 

The next screen information becomes the next root vector to learn the state + action = next state relationship with a deductive model.

In this experiment, Deduction receives the current state information and action information as input and predicts the root vector containing the next state information.

It is similar to receiving action and performing a step in the environment, and the next state is generated.

\paragraph{Discriminator}
In the discriminator process, Learn with a network that classifies whether the current state is an end state or not.
Discriminator in this experiment means whether the current state is the final target endpoint whenever the agent acts. 
Therefore, Discriminator was learned using done (whether the game was over or not).

\subsubsection{ Imagine Result : Model 1 }

\begin{figure}[h]
\centering
{\includegraphics[width=0.80\textwidth]{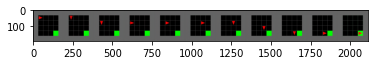}}
\caption{ A Generated Sample }
\label{fig:long-term}
\vspace{-10px}
\end{figure}
The learned network no longer generates data from the environment, but the following simulations are possible.
this figure is a generated sample and there are four characteristics. 

(image generation by state number and root vectors (memory), Continuous scenes(deduction), the shortest path(reinforcement learning), When is the end state?(Discriminator))

\paragraph{Discuss : How do we think creatively? }
\begin{equation}
G \oplus G \to G^{c} \cap G'
\label{eqn:grouptheory1}
\end{equation}
Creative samples appear when they are moved to other set of elements without being closed to any operation of the elements. 

If the element move from the current set $G$ to $G^{c}$, It moves away from the existing knowledge and reaches a different set space($G'$) and creates something new. And since it is a sample produced by deduction, it is theoretically valid.

I think we may create something new by designing the conditional expression, learning the discriminator, and generating a sample.

\section{Conclusion}
We are designing an agent model that behaves similarly to the human brain by combining various networks developed to date. The purpose of this study is as follows. "Let’s create a brain that thinks like humans in a similar environment to human life". This paper is part of a series. The next paper is ~.


\bibliographystyle{splncs04}  
\bibliography{main}  

\end{document}